\pdfoutput=1
\def\year{2022}\relax
\documentclass[letterpaper]{article} 
\usepackage{aaai22}  
\usepackage{times}  
\usepackage{helvet}  
\usepackage{courier}  
\usepackage[hyphens]{url}  
\usepackage{graphicx} 
\usepackage{natbib}  
\usepackage{caption} 
\DeclareCaptionStyle{ruled}{labelfont=normalfont,labelsep=colon,strut=off} 
\usepackage{algorithm}
\usepackage{algorithmic}

\usepackage[T1]{fontenc}
\usepackage{amsmath}
\usepackage{amssymb}
\DeclareMathOperator*{\argmax}{argmax}
\DeclareMathOperator*{\argmin}{argmin}
\usepackage{bm}
\usepackage{multirow}

\usepackage{booktabs}
\usepackage{newfloat}
\usepackage{listings}
\floatstyle{ruled}
\newfloat{listing}{tb}{lst}{}
\floatname{listing}{Listing}
\title{Robust Depth Completion with Uncertainty-Driven Loss Functions}
\author{
    Yufan Zhu,\textsuperscript{\rm 1} 
    Weisheng Dong,\textsuperscript{\rm 1}\thanks{Corresponding author(wsdong@mail.xidian.edu.cn).} 
    Leida Li,\textsuperscript{\rm 1} 
    Jinjian Wu,\textsuperscript{\rm 1} 
    Xin Li,\textsuperscript{\rm 2}  
    Guangming Shi\textsuperscript{\rm 1} 
    
}
\affiliations{
    \textsuperscript{\rm 1} School of Artificial Intelligence, Xidian University, Xi’an 710071, China\\
    \textsuperscript{\rm 2} Lane Dep. of CSEE, West Virginia University, Morgantown WV 26506, USA\\
    zhyf70695@163.com,
    wsdong@mail.xidian.edu.cn,
    ldli@xidian.edu.cn,
    jinjian.wu@mail.xidian.edu.cn,
    xin.li@mail.wvu.edu,
    gmshi@xidian.edu.cn

}

\begin{document}

\maketitle

\begin{abstract}
Recovering a dense depth image from sparse LiDAR scans is a challenging task. Despite the popularity of color-guided methods for sparse-to-dense depth completion, they treated pixels equally during optimization, ignoring the uneven distribution characteristics in the sparse depth map and the accumulated outliers in the synthesized ground truth. In this work, we introduce uncertainty-driven loss functions to improve the robustness of depth completion and handle the uncertainty in depth completion. Specifically, we propose an explicit uncertainty formulation for robust depth completion with Jeffrey's prior. A parametric uncertain-driven loss is introduced and translated to new loss functions that are robust to noisy or missing data. Meanwhile, we propose a multiscale joint prediction model that can simultaneously predict depth and uncertainty maps. The estimated uncertainty map is also used to perform adaptive prediction on the pixels with high uncertainty, leading to a residual map for refining the completion results. Our method has been tested on KITTI Depth Completion Benchmark and achieved the state-of-the-art robustness performance in terms of MAE, IMAE, and IRMSE metrics.
\end{abstract}

\section{Introduction}
\par
\begin{figure}[tb]
\centering
\includegraphics[width=0.45\textwidth]{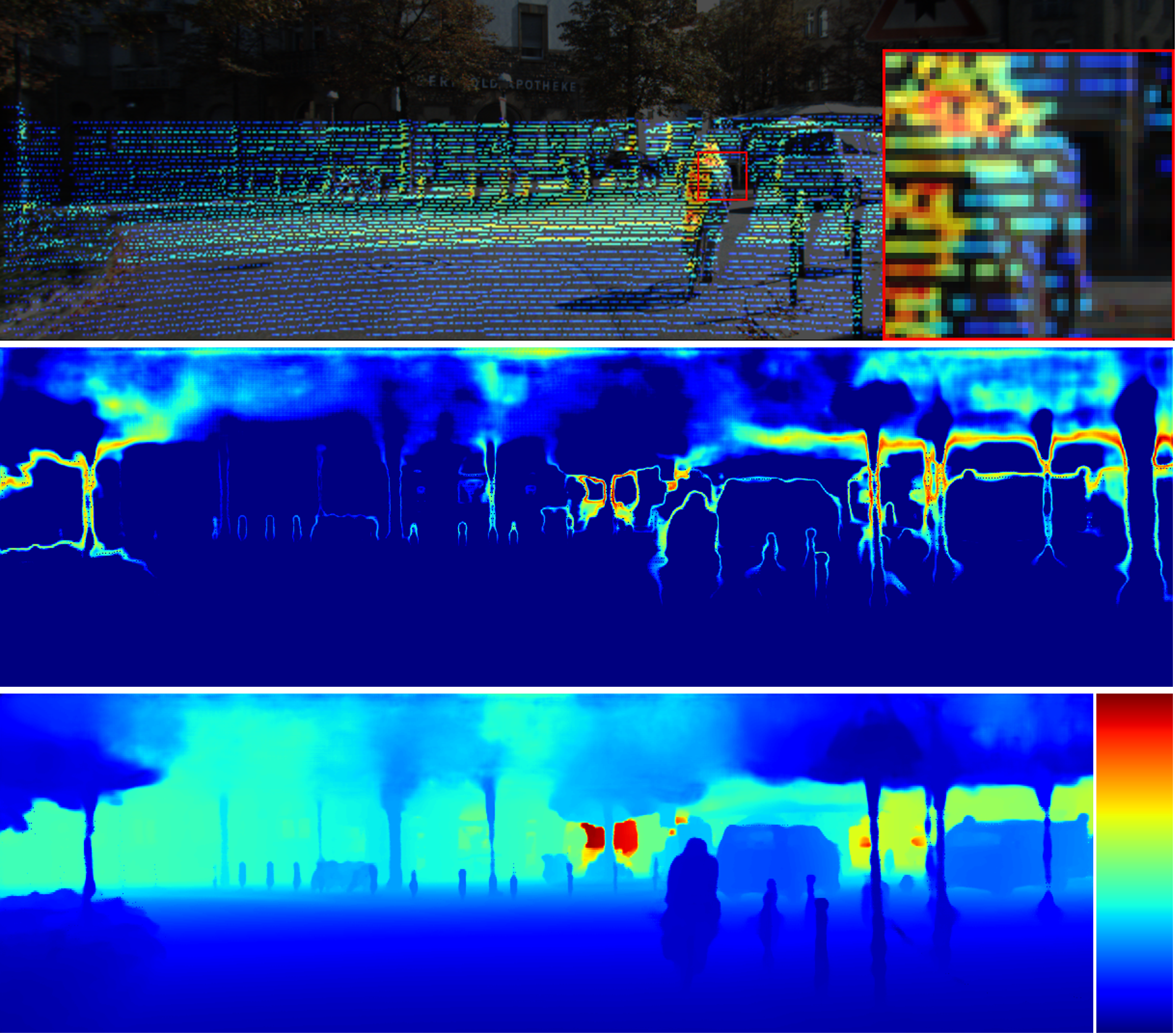} 
\caption{\textbf{Depth completion with uncertainty prediction.} \textbf{Top:} density distribution of a raw Lidar scans with a zoom-in part. Color is proportional to the local sampling density (blue to red is low to high density). To facilitate visual inspection, we have superimposed it on the RGB scene. \textbf{Middle:} the uncertainty map predicted by our method (note that high uncertainty is often associated with depth discontinuities). \textbf{Bottom:} the completed depth image with our method.}
\label{Fig:DepthCompletion}
\end{figure}

 Depth sensing has become increasingly important to a variety of 3D vision tasks, including human-computer interaction \cite{newcombe2011kinectfusion}, 3D mapping \cite{zhang2014loam}, and autonomous driving \cite{wang2019pseudo}. Depending on the application, 3D sensing of indoor and outdoor environments often faces different technical barriers. For indoor environments, 3D sensing such as Microsoft Kinect and Intel RealSense has become widely affordable, but often suffer from the problem of missing pixels in the presence of shiny/transparent surfaces or inappropriate camera distance \cite{huang2019indoor}. For outdoor environments, LiDAR is the most popular sensor for acquiring depth information. In addition to the high cost, a fundamental limitation with existing LiDAR sensors is the sparsity of depth measurements. Accordingly, so-called sparse-to-dense completion (a.k.a. depth completion) has been widely studied in the literature (e.g., DeepLiDAR \cite{qiu2019deeplidar}, GuideNet \cite{tang2020learning}, Adaptive Context-aware Multi-modal Network \cite{zhao2020adaptive}, Non-Local spatial propagation network \cite{park2020non}). Unlike image super-resolution (SR) \cite{zhang2018image}, the sparse depth map is the degradation (with \textbf{non-uniform} downsampling) of distance projection; while the low resolution (LR) RGB image is the \textbf{spatially uniform} sampling, which makes the problem of depth completion unique. For instance, we have to handle a significant portion of missing data, as shown in Fig.~\ref{Fig:DepthCompletion}(Top). Traditional full convolution network (FCN) models for SR methods (e.g., RCAN \cite{zhang2018image}, etc.) can not be directly applied to sparse depth maps due to the non-uniform sampling of sparse depth maps as well as the absence of high resolution (HR) RGB images as the ground truth (GT). 

\begin{figure}[t]
\centering
\includegraphics[width=0.48\textwidth]{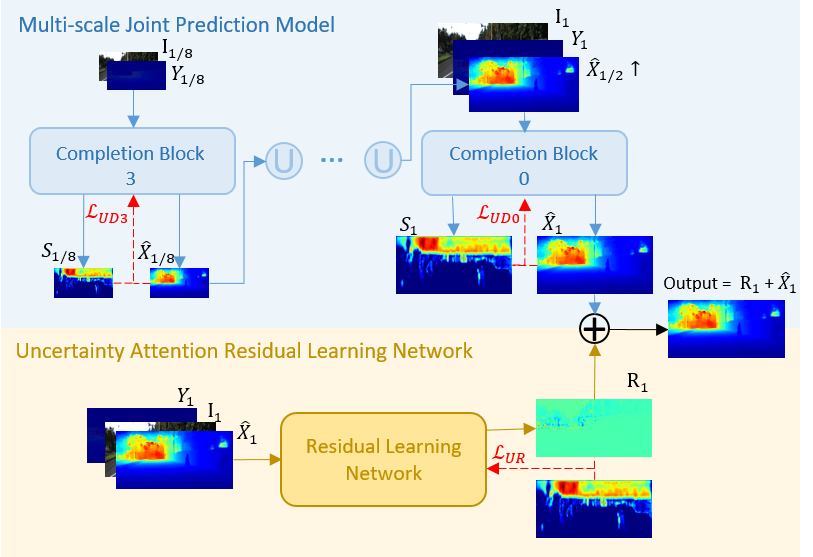} 
\caption{\textbf{Overview of our Method.} \textbf{Top}: we jointly predict the uncertainty maps and dense depth images using the Multi-scale Joint Prediction Model. \textbf{Bottom}: Uncertainty Attention Residual Learning Network is used to refine the prediction for pixels of high uncertainty.}
\label{Fig:MainTr}
\end{figure}

\par Two kinds of uncertainty were introduced in Bayesian deep learning for depth regression and segmentation \cite{kendall2017uncertainties}. Aleatoric uncertainty captures noise inherent in the observations, and the epistemic uncertainty explains the model uncertainty. In depth completion, aleatoric uncertainty captures noise inherent in raw LiDAR data, which is sparse and unevenly distributed. For example, LiDAR scans the surrounding environment at equally divided angles, resulting in an uneven distribution of depth samples\cite{uhrig2017sparsity}, as shown in Fig.~\ref{Fig:DepthCompletion}. The boundary area of objects and the ultra long-distance areas can hardly be scanned by LiDAR. Such an uneven distribution often leads to the poor prediction of areas with sparse samples. Moreover, the ground truth of KITTI dataset \cite{geiger2013vision} is synthesized by accumulating 11 laser scans and removing outliers by only comparing the depth results in stereo images. Therefore, many outliers will accumulate in semi-sparse GT depth images during synthesis.

\par In this paper, we introduce the uncertainty-driven loss function to address the uncertainty issue in sparse depth maps and solve the problem of depth completion more effectively. We propose a joint estimation method to simultaneously predict the missing depth values and their uncertainties under a probabilistic framework. By introducing Jeffrey's prior \cite{figueiredo2001adaptive} to the model, we obtain a new loss function robust to noisy LiDAR data. Inspired by the success of multiscale methods \cite{shaham2019singan, nah2017deep}), we have carefully designed a multiscale joint prediction network to concurrently estimate the depth and uncertainty maps in a coarse-to-fine manner. The extension of uncertainty-driven loss functions is further complemented by an adaptive prediction of the pixels with high uncertainty, leading to a residual map for refining the depth completion results. The main technical contributions of this work can be summarized as:
\begin{itemize}
\item Uncertainty-based deep learning framework for depth completion. For the first time, we propose to develop a more fundamental understanding of aleatoric uncertainty on LiDAR scans for the task of sparse depth completion. 
\item Uncertainty modeling and estimation. We propose a Multiscale Joint Prediction Model to simultaneously estimate depth and uncertainty maps. The introduction of Jeffrey’s prior and multiscale extension both contribute to the improved robustness of our depth completion approach.
\item Uncertainty attention residual learning. The aleatoric uncertainty-driven method \cite{kendall2017uncertainties} was further expanded by residual learning. A new uncertainty-attention network is developed to refine the predicted dense depth with high uncertainty measures.
\item Our method has been trained and tested on the popular KITTI benchmark \cite{geiger2013vision}. It has achieved the top-ranked performance in terms of MAE, IMAE, and IRMSE metrics among all published papers, which justifies the effectiveness of the proposed approach. 
\end{itemize}
\section{Related Work}
\noindent \textbf {Depth estimation from a single image}. Depth estimation from a single image \cite{eigen2014depth} has been extensively studied in the literature. A deep structured learning scheme was proposed in \cite{liu2015deep,long2015fully} to learn the unary and pairwise potentials of continuous conditional random field (CRF) by a unified deep convolutional neural network (CNN) framework. This line of research was extended into single-image depth estimation in the wild \cite{chen2016single} - i.e., recovering the depth from a single image taken in unconstrained settings. More recently, a spacing-increasing discretization (SID) strategy was introduced in \cite{fu2018deep} to discretize depth values and recast depth estimation as an ordinal regression problem.

\begin{figure*}[tb]
\centering
\includegraphics[width=0.95\textwidth]{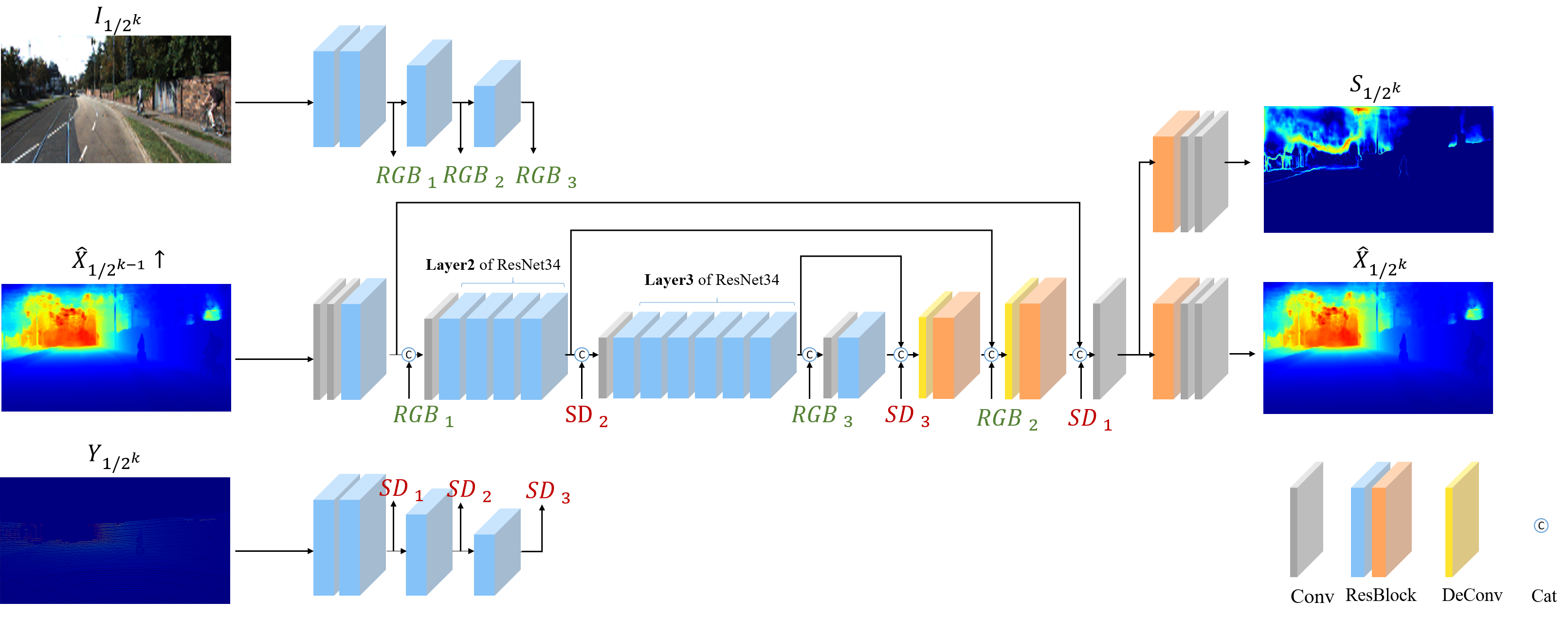} 
\caption{\textbf{A Depth Completion Block in Multi-scale Joint Prediction Network.} Input the up-sampling results of the previous module to generate fine-dense depth images by fusing with RGB and sparse depth.}
\label{Fig:CompletionBlock}
\end{figure*}

\noindent \textbf{Model-based sparse-to-dense depth completion methods}. Early methods of estimating dense measurements from sparse ones have been considered under the framework of compressed sensing. For example, \cite{hawe2011dense} shows disparity maps can be reconstructed from only a small set of reliable support points based on the compressive sensing principle. In \cite{liu2015depth}, a combined dictionary of wavelets and contourlets proved to improve the reconstruction quality of disparity maps. In \cite{ku2018defense}, a simple and fast method was developed to complete a sparse depth map using basic image processing operations only.       

\noindent \textbf{Deep learning based depth completion}. By designing the U-net and fusing the aligned color image, self-supervised approaches such as \cite{ma2019self}  have achieved significant improvement over traditional model-based methods. In \cite{van2019sparse}, a new framework was proposed to extract both global and local information from two different networks during depth completion. More recently, \cite{tang2019learning} introduced the guided convolution module into a network architecture to fuse color information and developed spatially separable convolution to reduce computational complexity. There are also methods\cite{qiu2019deeplidar,chen2019learning} using auxiliary variables to assist the task of depth completion. In \cite{qiu2019deeplidar}, surface normals are estimated from color images and imposed as a geometric constraint for depth completion. In \cite{chen2019learning}, sparse depth maps are converted to 3D point clouds, then both 2D and 3D features are exploited to improve the performance of depth completion.

\noindent \textbf{Uncertainty estimation.} Early work\cite{mackay1992practical,neal2012bayesian} like Bayesian neural networks (BNNs) has been designed to obtain uncertainty estimates. In the work\cite{kendall2017uncertainties}, the aleatoric uncertainty from the noise in the observations and the epistemic uncertainty explaining the model uncertainty were studied in a joint framework in the tasks like pixel-wise depth regression and semantic segmentation. And Recently, more methods \cite{qiu2019deeplidar,van2019sparse,xu2019depth} promoted the prediction of Uncertainty to filter out noisy predictions within the network.\cite{van2019sparse,xu2019depth} predicts the confidence(just like uncertainty) to fuse the completed depth differently, like local and global networks. \cite{qiu2019deeplidar} learned confidence mask to represent the noisy depth of occluded regions. More recently, researchers in \cite{chang2020data} have investigated the data uncertainty with estimated mean and variance in face recognition. \cite{wong2019bilateral,wong2021adaptive} developed adaptive regularization and data fidelity to handle hard boundary prediction in unsupervised depth completion.

\section{Approach}
\noindent We first propose a probabilistic method for joint training of depth and uncertainty maps. Then we present how to implement our proposed loss functions in a multiscale network to predict the depth and uncertainty maps from coarse to fine, which is named the Uncertainty-based Multi-scale Joint Prediction Model. Finally, we show how to use the predicted uncertainty map to improve the result of depth completion by an Uncertainty Attention Residual Learning Network. 

\begin{figure*}[tb]
\centering
\includegraphics[width=0.95\textwidth]{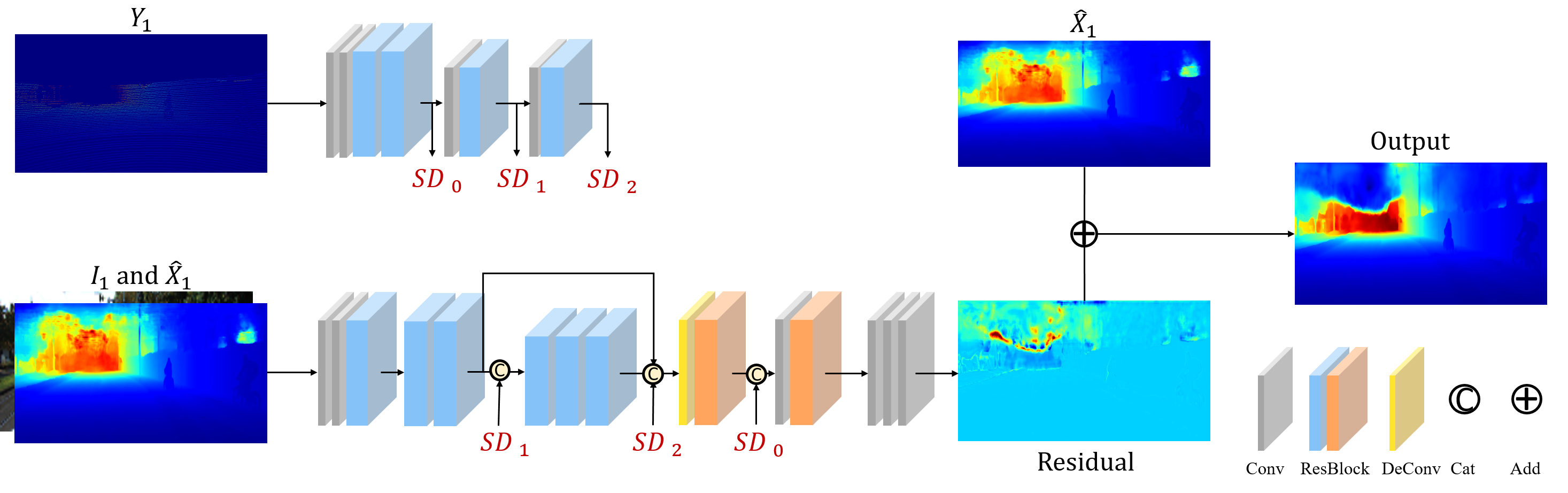} 
\caption{\textbf{Uncertainty Attention Residual Learning Network}.}
\label{Fig:ResdualBlock}
\end{figure*}

\subsection{Uncertainty-Driven Loss Function}
\label{section:PJT Method}
\noindent In low-level vision tasks, Bayesian deep learning offers a principled framework for taking uncertainty into account \cite{kendall2017uncertainties}. In Bayesian modeling, there are primarily two types of uncertainty: {\em aleatoric} and {\em epistemic}. Aleatoric uncertainty capturing noise inherent in the observations can be further categorized into two classes: homoscedastic and heteroscedastic. Under the context of depth completion, heteroscedastic uncertainty is especially important due to the physical limitations of LiDAR sensors. For example, LiDAR scans the surrounding environment at equally divided angles, resulting in an uneven distribution of depth images \cite{uhrig2017sparsity}; such an uneven distribution often leads to varying difficulties in different densities areas.

In previous depth completion methods \cite{ma2019self,qiu2019deeplidar,chen2019learning}, the MSE Loss was used to average the errors across all pixels. Low-density areas (arising from non-uniform sampling) and outliers often cause the network to over-emphasize these areas (i.e., overfitting).
Inspired by the recent work of uncertainty modeling \cite{lee2019gram}, we consider a parametric approach of quantifying the uncertainty in a depth map by its variance field $\Sigma$. 
First, we define the sparse depth image as $Y$, the corresponding dense depth image (GT) as $X$, and the process of generating dense depth images based on the deep learning network is $X = F(Y)$, where we expect the predicted $\hat{X}$ to approximate $X$. Thus, the procession of depth completion can be expressed as maximizing the posterior probability $P\left(X|Y\right)$. By introducing the uncertainty measure $\Sigma$ ($\sigma$ for a pixel), we can decompose the joint posterior probability into the product of marginals: 
\begin{equation}
\small
\begin{split}
     P\left(X,\Sigma|Y \right) =& P\left(\Sigma|Y\right)P\left(X|\Sigma,Y \right) \\
    =& \prod p\left(\sigma_i|y_i\right)p\left(x|\sigma_i,y_i\right)
\label{eq:1}
\end{split}
\end{equation}
where $x_i$, $\sigma_i$, $y_i$ is the pixel-wise element of $X$, $\Sigma$, and $Y$. For the likelihood of uncertainty map $p\left(\sigma_i|y_i\right)$, we model it with the Jeffrey's prior $P\left(\sigma_i|y_i\right) \approx \frac{1}{\sigma_i}$ \cite{figueiredo2001adaptive} based on the intuition of the sparsity on uncertainty map. For the likelihood term, $p\left(x_i|\sigma_i,y_i \right)$ can be modeled by a Gaussian distribution observing $\hat{x}_i = F(y_i) \sim N\left(x_i,\sigma_i\right)$: 

\begin{equation}
\small
p\left(x_i|\sigma_i,y_i \right) \approx \frac{1}{\sqrt{2\pi}\sigma_i}\exp(-\frac{\left(\hat{x}_i - x_i \right)^2}{2\sigma_i^2})
\label{eq:2}
\end{equation}
where $\hat{x}_i$ is a pixel of $\hat{X}$. Therefore, we obtain the following MAP problem:  

\begin{equation}
\small
\begin{split}
& \max \sum \left(\log p\left(\sigma_i|y_i\right) + \log p\left(x_i|\sigma_i,y_i \right)\right) \\
=& \argmax_{\hat{x}_i, \sigma_i} \sum \left(-2\log \sigma_i - \frac{\left(\hat{x}_i - x_i \right)^2}{2\sigma_i^2} - \frac{1}{2}\log{2\pi} \right) \\
=& \argmin_{\hat{x}_i, \sigma_i} \sum \left(4\log \sigma_i + \frac{\left(\hat{x}_i - x_i \right)^2}{\sigma_i^2}\right) \\
=& \argmin_{\hat{x}_i, s_i} \sum \left(e^{-s_i} \left(\hat{x}_i - x_i \right)^2 + 2s_i\right)
\end{split}
\label{eq:3}
\end{equation}
where $s_i = 2 \log \sigma_i $, $\sigma_i^2 = e^{s_i}$. Such formulation of uncertainty modeling can be translated to the design of a new loss function. Conceptually similar to the previous work \cite{kendall2017uncertainties}, we can build the final optimized loss, named Uncertainty-Depth Joint Optimization Loss Function as follows:

\begin{equation}
\small
\mathcal{L}_{UD} = \frac{1}{N} \sum \left(e^{-s_i} \left(\hat{x}_i - x_i \right)^2 + 2s_i\right)
\label{eq:4}
\end{equation}

From the formula, we conclude that the first term will reduce the joint loss of pixels with large differences between the prediction and ground truth $\left(\hat{x}_i - x_i \right)^2$. During the process of optimization, the optimizer may increase the uncertainty values so large that the penalty term $e^{-s_i}$ eventually approaches zero. To balance the first term, the second term limits the growth of uncertainty $s_i$ as a regularization term. As a consequence of balancing, the network will control the contribution of high-uncertainty regions to the joint loss function, instead of overfitting these regions. In summary, Eq. \eqref{eq:4} makes the overall network focused more on easy-to-predict regions by attenuating hard-to-predict regions to strike an improved tradeoff.

\subsection{Multiscale Joint Prediction Model}
\label{section:MS Joint Prediction Model}
\noindent We present how to implement the joint uncertainty-depth optimization model. Our network implementation contains two basic modules: the Multi-scale joint prediction network as shown in Fig. \ref{Fig:MainTr}(Top), the Sparse-to-dense Basic Completion Block as shown in Fig. \ref{Fig:CompletionBlock}.

\begin{figure*}[tb]
\centering
\includegraphics[width=0.95\textwidth]{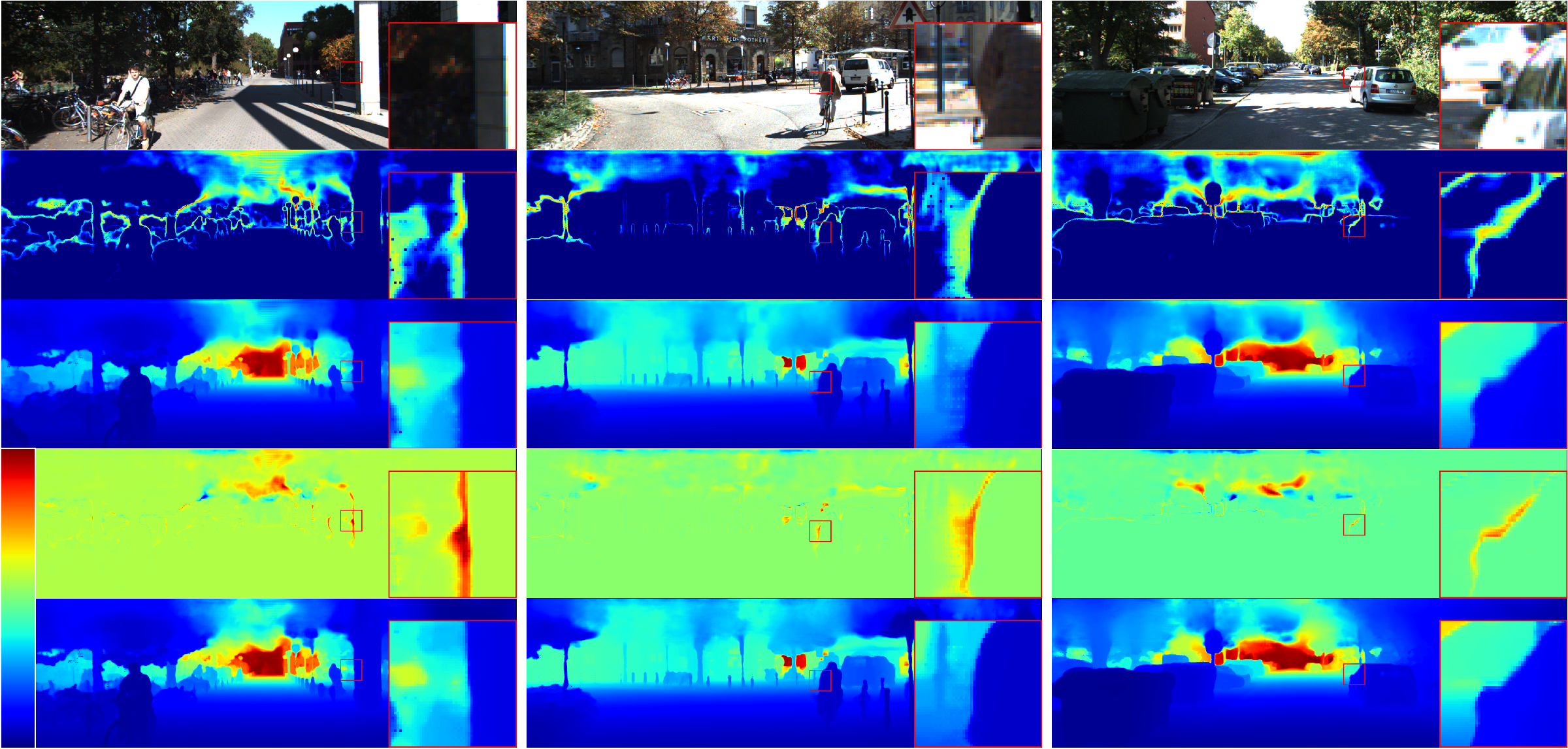} 
\caption{\textbf{Demonstration of depth completion results after each stage.} \textbf{ First row:} the color images. \textbf{Second row:} the uncertainty maps. \textbf{Third row:} the dense depth images generated by Multi-scale Joint Prediction Model. \textbf{Fourth row:} the residual maps generated by Uncertainty Attention Residual Learning Network. \textbf{Fifth row:} the final output images. }
\label{Fig:imgComparison}
\end{figure*}

\noindent \textbf{Multiscale Network Structure}. Similar multiscale architecture has been shown effective for various low-level vision tasks such as image synthesis \cite{shaham2019singan} and image deblurring \cite{nah2017deep}. The multiscale module starts with a pair of down-sampled color image $I_{1/2^k}$(Interpolation) and depth map $Y_{1/2^k}$(Max-Pooling), where $k$ is the downsampling factor. To handle sparsity, we use a relatively large-size convolution kernel to extract and propagate spatial features from the coarse to fine scale. Throughout a series of upsampling operations after completion, the size of the convolution kernel is kept constant, but the size of image patches (receptive field) decreases relatively to rich more details. Unlike previous works, we have found that an interpolation upsampling layer in the pixel domain between completion blocks, as shown in Fig.~\ref{Fig:MainTr}, is preferred for coarse-to-fine progression over an upconvolution layer, which has strong local constraints than the deconvolution process (often used for superresolution in the feature domain \cite{sajjadi2017enhancenet}). Meanwhile, such a multiscale framework allows us to selectively supervise the output(depth and uncertainty) of varying patch sizes and adaptively calculate the loss function for performance optimization. To optimize the uncertainty and depth values at all scales at the same time, we define the following weighted multi-objective optimization function. 
\begin{equation}
\begin{split}
\mathcal{L}_{StepOne} =\sum \omega_{k}\mathcal{L}_{UDk} 
\label{eq:5}
\end{split}
\end{equation}
where $\omega_k$ and $\mathcal{L}_{UDk}$ denote the weighting coefficient and loss term at the $k$-th scale, respectively.

\noindent \textbf{Sparse-to-dense Basic Completion Block}. The completion block's backbone is constructed based on the well-known U-net that has shown effectiveness in image segmentation tasks \cite{ronneberger2015u}. In our construction, the backbone of the model consists of two branches, each of which will output the depth image and uncertainty map respectively. Similar architecture has also been used in self-supervised depth completion \cite{ma2019self} where it takes the layers of original Resnet34 \cite{he2016deep} as the encoder and some deconvolution layers of \cite{zeiler2011adaptive} as the decoder. By contrast, our design reflects the unique strategy of fusing color and depth information in an alternating manner, as marked by the bars with different colors and heights in Fig. \ref{Fig:CompletionBlock}. Additionally, we have added some skip connections to the residual modules to further facilitate the information flow between deconvolution operations \cite{van2019sparse}.

\begin{figure*}[tb]
\centering
\includegraphics[width=0.95\textwidth]{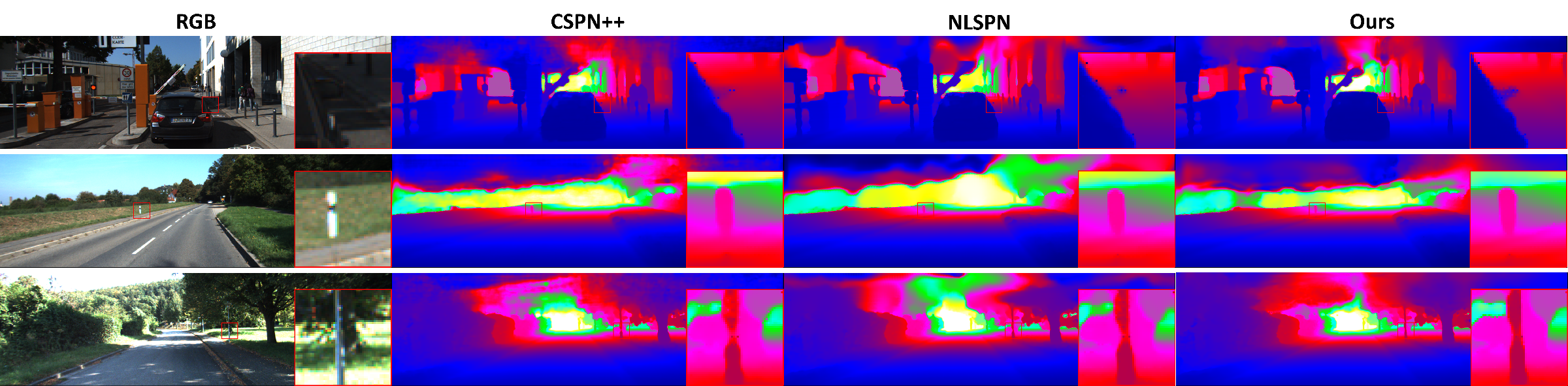} 
\caption{\textbf{Visual quality comparison on KITTI Test Benchmark\cite{uhrig2017sparsity}.} \textbf{Left to right:} RGB, CSPN++\cite{cheng2020cspn++}, NLSPN\cite{park2020non}, Our results.}
\label{Fig:SOTA}
\end{figure*}

\subsection{Uncertainty Attention Residual Learning Network}
\noindent Through the previous analysis, we know that the model has achieved an improvement in the overall recovery performance by alleviating those difficult-to-recover areas (regions with great uncertainty). From the predicted uncertainty map in Fig. \ref{Fig:imgComparison}, we can observe that the edge regions of the object and the regions with larger depth values often have larger uncertain values. Note that the optimization constraints of these regions are relatively relaxed in the first step (Multiscale Joint Prediction). The new insight behind our residual learning network in the second step is to use the estimated uncertainty map to guide the procedure of depth completion {\em refinement}. In other words, with the knowledge about the distribution of high-uncertainty regions, we hope to tailor the process of optimization for these special regions to achieve an even better completion result.

Our key idea is to predict the refinement map $R$ for $\hat{X}_1$, only for the pixels that are uncertain in the first step. Note that the predicted uncertainty map $S_1$ is used to give higher weight to the loss in the regions of high uncertainty. Based on the above reasoning, the loss function associated with residue learning can be formed by:
\begin{equation}
\small
\begin{aligned}
&\mathcal{L}_{UR} = \frac{1}{N} \sum s_i |\left( x_i - \hat{x}_i\right) - r_i| \\
&\mathcal{L}^{2}_{UR} = \frac{1}{N} \sum s_i \left(\left( x_i - \hat{x}_i\right) - r_i\right)^{2}
\end{aligned}
\label{eq:6}
\end{equation} 
where $r_i$ is the pixel of predicted residual $R$, $\hat{x}_i$ is the depth output of Multiscale Joint Prediction Network and we will use a mixture of L1 and L2 forms to build the epoch-dependent balanced loss function next. The structure of the network is a simplified U-net, which takes RGB images and $\hat{X}_1$ as input, incorporates the information of the sparse depth in the prediction process, and finally outputs the residual image. The final output of depth completion is $R+\hat{X}_1$, as shown in Fig.~\ref{Fig:ResdualBlock}.

It is worth mentioning that the optimization of different objective metrics for depth completion often has conflicting objectives. For example, the objective of minimizing RMSE is often inconsistent with that of minimizing MAE (the former is more sensitive to outliers than the latter). Note that RMSE is equivalent to the square root of L2 loss and MAE is equivalent to the L1 loss; they respectively characterize different performance metrics for the task of depth completion. A compromised solution is to formulate a balanced loss function between different metrics. In our current implementation, we have used the following epoch-dependent balanced loss function for the residual learning network: 

\begin{equation}
\small
\mathcal{L}_{StepTwo}=\mathcal{L}_{URB}=\left\{
\begin{aligned}
& \mathcal{L}_{UR},& N_{epoch}~is~even  \\
& \frac{1}{2}\left(\mathcal{L}_{UR} + \mathcal{L}^{2}_{UR}\right), & else\\
\end{aligned}
\right.
\end{equation}
In summary, this epoch-dependent loss function aims at better balancing the RMSE and MAE metrics by adaptively combining L1 and L2 losses. The benefit of using a hybrid yet balanced loss function is shown in Table~\ref{Tab:BalanceLoss}.  
\begin{table}[ht]
\small
\centering
\begin{tabular}{@{}lllll@{}}
\toprule
\textbf{Methods}      & \textbf{MAE}    & \textbf{IMAE}                                   & \textbf{RMSE} & \textbf{IRMSE} \\ \midrule
NLSPN                 & 199.59          & 0.84                                            & 741.68        & 1.99           \\
GuideNet             & 218.83          & 0.99                                            & 736.24        & 2.55          \\
CSPN++                & 209.28          & 0.90                                            & 743.69        & 2.07           \\
DeepLiDAR             & 226.50          &1.15                                             & 758.38        & 2.56           \\
Sparse-to-Dense (gd)  & 249.95          & 1.21                                            & 814.73        & 2.80           \\
RGB$\_$guide$\&$certainty & 215.02          & 0.93                                          & 772.87        & 2.19           \\
Ours(with $L_{URB}$)                  & 198.09 & 0.85                                            & 751.59        & \textbf{1.98}  \\
Ours(with $L_{UR}$)                   & \textbf{190.88} & \textbf{0.83}                                            & 795.43        & \textbf{1.98}  \\

\bottomrule
\end{tabular}
\caption{\textbf{Comparison with other SOTA methods on KITTI Test benchmark.}}
\label{Tab:SOTA}
\end{table}
\section{Experiments}

\subsection{Experimental Settings}
\noindent \textbf{Dataset}. The KITTI depth completion benchmark\cite{uhrig2017sparsity} has 86898 Lidar frames for training, 1000 frames for validation, and 1000 frames for testing. Each frame has one depth map from the LiDAR sensor and one aligned color image from the visible spectrum camera. 

\noindent \textbf{Evaluation Metrics}. Root Mean Square Error (RMSE), Mean Absolute Error (MAE), Inverse RMSE (IRMSE), and inverse MAE (IMAE). 

\noindent \textbf{Parameter Settings}. Our training is implemented by Pytorch with 5 NVIDIA GTX2080Ti GPUs and set batch-size to 5. In our current implementation, we have used ADAM \cite{kingma2014adam} as the optimization algorithm. We have set the learning rate to $1 \times 10^{-4}$ when we train our multiscale joint prediction model and $2 \times 10^{-4}$ when training uncertainty attention residual learning model. The acceleration of learning rate will decline as the epoch increases. The other parameters are all the same with $(\beta_1$, $\beta_2)$ = $(0.9,0.999)$, $eps = 1 \times 10^{-8}$ and $Weight\_decay = 0$. 

\noindent \textbf{Mask Loss.} Since the groundtruth is semi-dense, a binary mask is applied to the loss function only accounting for the valid depth points.



\subsection{Performance Comparison}

\noindent \textbf{Comparison with SOTA methods.} We have compared our method with the SOTA methods on the KITTI TEST benchmark. Our method has achieved highly competitive results in terms of all metrics. As shown in Table.\ref{Tab:SOTA},  it surpasses all other competing methods on MAE, IMAE, and IRMSE metrics, which demonstrates the superiority of our method.

\begin{table}[tb]
\centering
\small
\begin{tabular}{@{}lllllll@{}}
\toprule
\textbf{Unc} & \textbf{Jef} & \textbf{Res} & \textbf{MAE} & \textbf{IMAE} & \textbf{RMSE} & \textbf{IRMSE} \\ \midrule
$\times$ & $\times$ & $\times$ & 211.01 & 0.92 & 792.34 & 2.185 \\
$\checkmark$ & $\times$ & $\times$ & 203.94 & 0.86 & 791.13 & 2.048 \\
$\checkmark$ & $\checkmark$ & $\times$ & 204.61 & 0.87 & 783.87 & 2.072 \\
$\checkmark$ & $\checkmark$ & $\checkmark$ & 186.90 & 0.81 & 833.67 & 2.069
 \\ \bottomrule
\end{tabular}
\caption{\textbf{Ablation study} of three key modules on Val set (uncertainty-driven loss, Jeffrey's prior, and residue learning).}
\label{Tab:modules}
\end{table}

\noindent We have also shown some qualitative visual comparison results on the test dataset of KITTI depth completion benchmark in Fig. \ref{Fig:SOTA}. Our results have clear boundaries and recover more details than other methods. For example, in the first row, our method is the only one capable of restoring the rearview mirror hidden in the dark background; in the third row, the vertical pole is better recovered by our method. 

\subsection{Ablation Studies and Analyses} We have tested the influence of several main modules on the experimental results. As shown in Table \ref{Tab:modules}, all three modules have jointly contributed to the performance improvement. However, the impact of different module varies on different quality metrics - e.g., MAE and IMAE are more consistent with each other, while RMSE is not. It can be verified that introduction of the Jeffrey prior greatly boosts the RMSE performance, which validates its effectiveness. 
\par We have also experimented on val set with the effect of different numbers of depth completion blocks (\textbf{NS} in the Table. \ref{Tab:Multi-scale}), and found NS = 4 reached the best metrics (3 out of 4) when the smallest downsampled image size is $\left(152,44\right)$(hard to be downsampled for U-net). 
In addition to using the sparse depth images as input correction information in Uncertainty Attention Residual Learning Network (Table. \ref{Tab:Input}), we have tested the effects of different input combinations on the results. It turns out that the combination of $\hat{X}_1$ and RGB image has achieved the best performance. 

\begin{table}[tb]
\small
\centering
\begin{tabular}{@{}lllll@{}}
\toprule
\textbf{NS} & \textbf{MAE} & \textbf{IMAE} & \textbf{RMSE} & \textbf{IRMSE} \\ \midrule
1           & 215.06       & 0.93         & 786.02        & 2.187          \\
2           & 205.95       & 0.88         & 778.05        & 2.110          \\
3           & 206.46       & 0.88         & 782.18        & 2.113          \\
4           & 204.61       & 0.87         & 783.87        & 2.072          \\ \bottomrule
\end{tabular}
\caption{\textbf{Ablation experiment} on the Number (NS) of S-to-D Basic Completion Block in Multiscale Joint Prediction.}
\label{Tab:Multi-scale}
\end{table}

\begin{figure}[tb]
\centering
\includegraphics[width=0.48\textwidth]{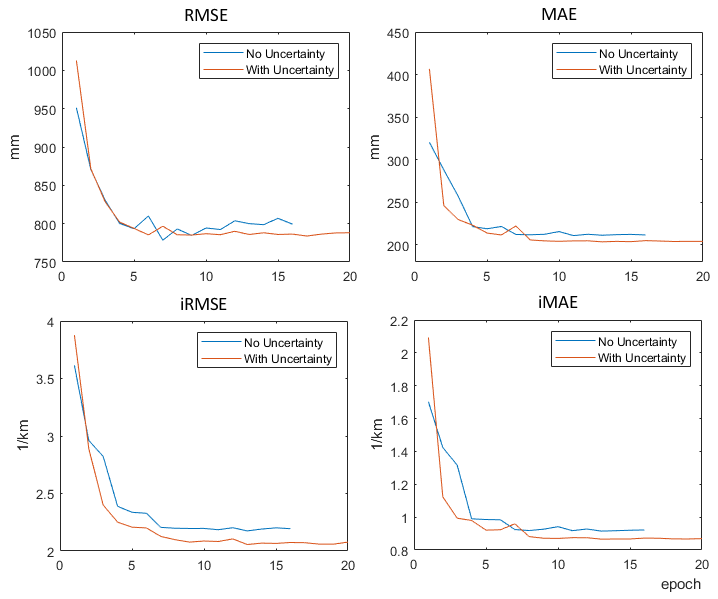} 
\caption{\textbf{Convergence analysis.} Training with uncertainty loss converges faster and reach lower bound than w/o. }
\label{Fig:Metrics}
\end{figure}

\begin{table}[tb]
\centering
\begin{tabular}{@{}llllll@{}}
\toprule
\textbf{$\hat{X}_1$}        & \textbf{$I_1$(RGB)} & \textbf{MAE} & \textbf{IMAE} & \textbf{RMSE} & \textbf{IRMSE} \\ \midrule
$\times$           & $\checkmark$        & 196.48       & 0.847         & 791.01        & 2.066          \\
$\checkmark$       & $\times$            & 187.31       & 0.809         & 834.33        & 2.037          \\
$\checkmark$       & $\checkmark$        & 186.90       & 0.807         & 833.67        & 2.069          \\ \bottomrule
\end{tabular}
\caption{\textbf{Ablation experiment} on the input of Uncertainty Attention Residual Learning module.}
\label{Tab:Input}
\end{table}

\begin{table}[ht]
\small
\centering
\begin{tabular}{@{}llllll@{}}
\toprule
\textbf{Residual} & \textbf{Loss}           & \textbf{MAE}    & \textbf{IMAE} & \textbf{RMSE}   & \textbf{IRMSE} \\ \midrule
$\times$        & $L_{UD}$      & 204.61 & 0.87 & 783.87 & 2.072  \\
$\checkmark$    & $L_{UR}$      & 186.90 & 0.81 & 833.67 & 2.069  \\
$\checkmark$    & $L_{URB}$     & 195.09 & 0.83 & 786.73 & 2.018  \\ \bottomrule
\end{tabular}
\caption{\textbf{Ablation experiment} on the loss function of Uncertainty Attention Residual Learning.}
\label{Tab:BalanceLoss}
\end{table}

\subsection{Convergence and Qualitative Analysis}

\noindent \textbf{Multiscale Joint Prediction Model.} This module will jointly predict the uncertainty map with dense depth image. From Table. \ref{Tab:modules} and Fig. \ref{Fig:Metrics},  we can observe that better results and robustness can be achieved by uncertainty-driven loss. By visually inspecting the uncertainty map in Fig. \ref{Fig:imgComparison}, we clearly see that the areas with high uncertainty are concentrated in the area of the object boundary (the depth drop is large) and the area with the higher depth value (distant road or open area). 
In Fig. \ref{Fig:Metrics}. Our empirical studies have shown that uncertainty-driven loss tends to produce a more chaotic start, but the optimization of the network will converge faster, and the objective metrics will stabilize around the optimal values after the convergence.
\begin{figure}[ht]
\centering
\includegraphics[width=0.40\textwidth]{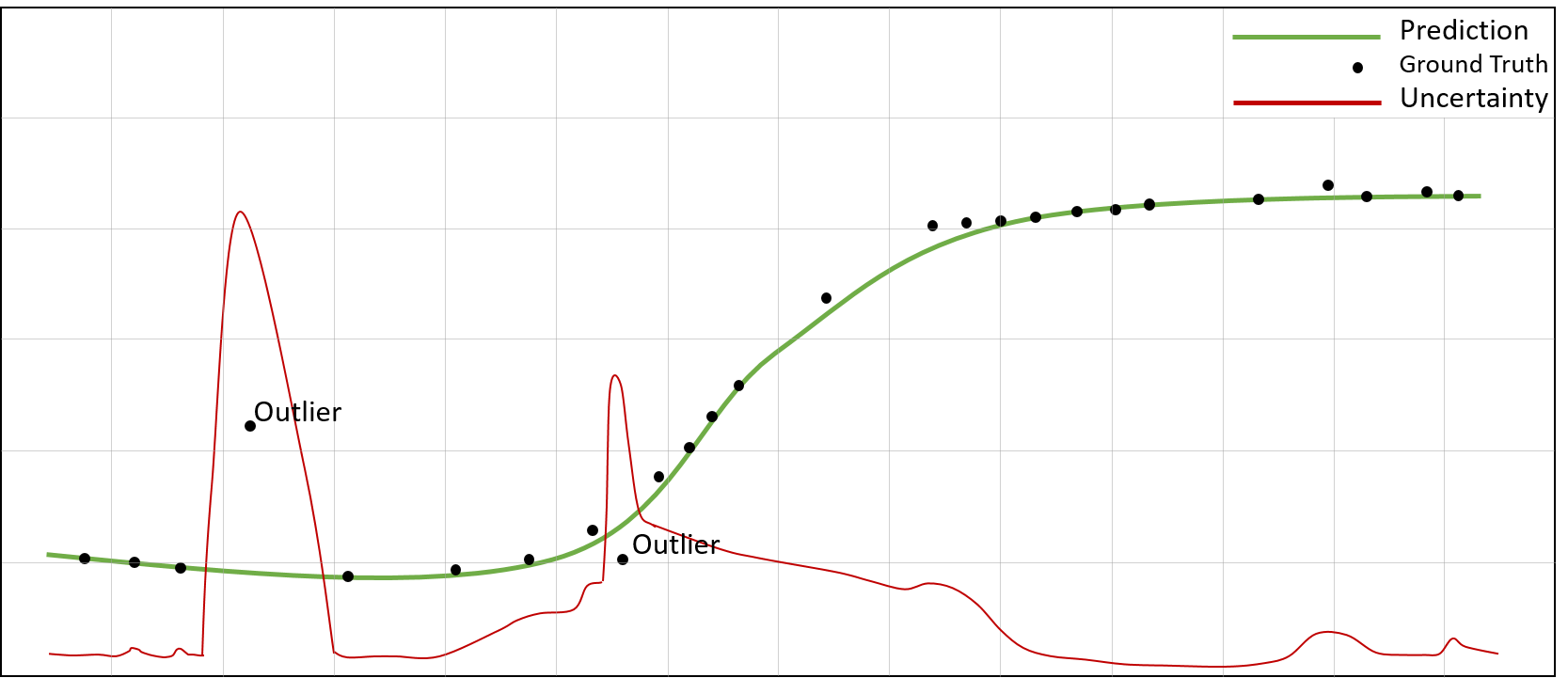} 
\caption{{\textbf{A qualitative example for the uncertainty in our method.} Areas with outliers, sparse, or large changes in depth values have higher uncertainty values, and these areas are weighted less during the joint training process.}}
\label{Fig:uncertainty_vision}
\end{figure}

\noindent \textbf{Uncertainty Attention Residual Learning Network.} We reuse the uncertainty map to focus on the part that contains more uncertainty in the multiscale joint prediction model. Table. \ref{Tab:modules} and \ref{Tab:BalanceLoss} verify how the prediction of the residual map boosts the performance. Visual inspection of the residual map in Fig. \ref{Fig:imgComparison} shows that most areas are close to zero (light blue) and high uncertainty areas are corrected (red is increasing and blue is decreasing). For a more intuitive explanation, please see a concrete example as shown in Fig. \ref{Fig:uncertainty_vision}. It clearly demonstrates that outliers are assigned high uncertainty values and therefore get attenuated by the joint training process.

\section{Conclusion}
\par In this paper, we introduce uncertainty into deep completion and propose how to improve the estimation for areas with high uncertainty. We propose a probabilistic Joint Training Method with Jeffrey's prior, consisting of the Multi-scale Joint Prediction Network and the Uncertainty Attention Residual Learning Network. Our method overcomes the difficulties caused by the uneven distribution and outliers in both LiDAR scanned depth images and synthesized semi-dense images. Extensive experimental results are reported to show that our method has better performance than existing top-rank published methods on the KITTI depth completion benchmark. Meantime, the computational complexity of our method is the lowest among the top methods - the actual running time of only 0.07s per frame can meet the requirement of real-time processing in practical applications. In addition, the related structure of uncertainty prediction can be removed totally after training, which will reduce network parameters further. In the future, we plan to study multimodal data fusion with LiDAR and other sensors. 
\section{Acknowledgments}
This work was supported in part by the National Key R\&D Program of China under Grant 2018AAA0101400 and the Natural Science Foundation of China under Grant 61991451, Grant 61632019, Grant 61621005, and Grant 61836008. Xin Li’s work is partially supported by the NSF under grants IIS-1951504 and OAC-1940855, the DoJ/NIJ under grant NIJ 2018-75-CX-0032, and the WV Higher Education Policy Commission Grant (HEPC.dsr.18.5).
\bibliography{aaai22}
\end{document}